\def\year{2019}\relax
\documentclass[letterpaper]{article} 
\usepackage{aaai19}  
\usepackage{times}  
\usepackage{helvet}  
\usepackage{courier}  
\usepackage{url}  
\usepackage{graphicx}  
\usepackage{amsfonts}
\usepackage{amsmath}
\usepackage{amssymb}
\usepackage{algorithm}
\usepackage{algorithmic}
\usepackage{subcaption}
\newcommand{\citet}[1]
{\citeauthor{#1}˜\shortcite{#1}}
\newcommand{\citep}{\cite}

\usepackage[usenames, dvipsnames]{color}
\DeclareMathOperator*{\argmin}{\arg\!\min}
\DeclareMathOperator*{\argmax}{\arg\!\max}
\begin{document}
%
\title{Inspiration Learning through Preferences}
\author{Nir Baram, Shie Mannor\\
Technion- Israel Institute of Tehnology\\
Haifa, 3200003\\
nirb@campus.technion.ac.il\\
shie@ee.technion.ac.il\\
}
\maketitle
\begin{abstract}
Current imitation learning techniques are too restrictive because they require the agent and expert to share the same action space. However, oftentimes agents that act differently from the expert can solve the task just as good. For example, a person lifting a box can be imitated by a ceiling mounted robot or a desktop-based robotic-arm. In both cases, the end goal of lifting the box is achieved, perhaps using different strategies.
We denote this setup as \textit{Inspiration Learning} - knowledge transfer between agents that operate in different action spaces.
Since state-action expert demonstrations can no longer be used, \textit{Inspiration learning} requires novel methods to guide the agent towards the end goal. In this work, we rely on ideas of Preferential based Reinforcement Learning (PbRL) to design Advantage Actor-Critic algorithms for solving inspiration learning tasks. Unlike classic actor-critic architectures, the critic we use consists of two parts: a) a state-value estimation as in common actor-critic algorithms and b) a single step reward function derived from an expert/agent classifier. We show that our method is capable of extending the current imitation framework to new horizons. This includes continuous-to-discrete action imitation, as well as primitive-to-macro action imitation. 
\end{abstract}

\noindent 

\section{Introduction}

Imitation Learning is an inter-discipline territory \cite{attia2018global}. What originally stemmed as a supervised learning problem \cite{pomerleau1991efficient} has been ever since promoted by members of the reinforcement learning community \cite{daume2009search}, \cite{ross2011no}.
Imitation learning has been successfully applied as an end-to-end solution \cite{ho2016generative}, or as a building block in more evolved engineering architectures \cite{silver2016mastering}.
Accommodating imitation learning concepts when training artificial agents is beneficial for several reasons. Techniques of such nature usually converge faster \cite{ross2010efficient} and with fewer unanticipated artifacts in the converged policy (a.k.a \textit{Reward hacking}) \cite{schaal1999imitation}. The merits of Imitation learning, together with current limitations of alternative reinforcement learning approaches \cite{rlblogpost} makes it a key component in the design of intelligent artificial agents.\\

\subsection{Current Imitation Techniques are too Restrictive}
Most commonly, imitation learning refers to the imitation of humans by robots or other artificial agents \cite{schaal1999imitation}.
Existing methods of imitation learning attempt to follow the expert’s policy directly. In other words, agents are trained to recover the state-action mapping induced by the expert. A fundamental underlying assumption of this approach is that the agent can at all act like the expert. Namely, that the expert and agent share the same action space. In the general case where the expert's and agent's action spaces are different, i.e. $\mathcal{A}_{expert} \neq \mathcal{A}_{agent}$, most, if not all existing approaches will fail because pure imitation is no longer feasible. Furthermore, the restriction that $\mathcal{A}_{expert}=\mathcal{A}_{agent}$ has far-reaching consequences when imitation is applied to real-world applications. In the context of robotics, it requires robots to have a humanoid structure, and in the context of self-driving cars, it favors the use of continuous-action agents since humans operate in this domain. As a result, the adoption of imitation learning in real-world applications has been severely hindered \cite{schaal2010learning}.

\subsection{Scalable Imitating Should be Action-Space Agnostic}
We focus on imitation problems where the state of the agent/expert is irrelevant for success. We refer to this type of problems as agent-agnostic tasks\footnote{Not all imitation tasks can be formulated in an agent-agnostic manner. There also exist tasks with an inherent dependence on the internal state of the agent. For example, teaching a robot to dance. We refer to this type of tasks as agent-dependent tasks.}. This setup covers most of the imitation tasks that come to mind\footnote{It is worth noting that oftentimes agent-dependent tasks can be formulated as agent-agnostic tasks. For instance, a walking task can be rephrased in an agent-agnostic manner by including the expert’s location (center of mass) in the state.}. Included here are object manipulation \cite{asfour2008imitation} and maneuvering tasks \cite{abbeel2004apprenticeship}. 
By definition, proper imitation in agent-agnostic tasks would aim to imitate the influence experts have on the environment rather than their explicit sequence of actions. In other words, scalable imitation should try to imitate the transition of environmental states induces by the expert, rather than its policy $\pi_E$. We denote such a \textit{loose} form of imitation as Inspiration Learning since the agent is free to craft new strategies, as long as their effect on the environment remains the same. Figuratively speaking, if a camera would be used to record expert demonstrations, then in the standard imitation approach it would be set to record the expert, while in the inspiration approach it would be set to record what the robot sees, i.e., the environment.
\subsection{Knowledge Transfer via State Transitions}
Transferring knowledge between an expert and an agent that do not share the same action space requires creative ways to evaluate whether the agent has learned how to carry out the task at hand or not. In this work, we try to address this challenge.
We argue that in the context of sequential decision-making problems, attending this question is possible by monitoring state transitions.
We turn to the celebrated actor-critic architecture and design a dyadic critic specifically for this task. The critic we use consists of two parts: 1) a state-value estimation part, as in common actor-critic algorithms and 2) a single-step reward function derived from an expert/agent classifier. 
This critic, which is oblivious to the action space of both players, is able to guide any agent toward behaviors that generate similar effects on the environment, analogous to that of the expert, even if the eventual policy is completely different from the one demonstrated by the expert.

\section{Related Work}
\label{sec:related-work}
In this section, we revisit key milestones in the field of imitation learning, starting from basic supervised approaches, and up to generative adversarial based imitation. Lastly, we briefly review the field of Preferential based Reinforcement Learning (PbRL), a concept that we believe can improve current imitation learning approaches.

\subsection{The Early Days of Imitation Learning}
Not surprisingly, the first attempts to solve imitation tasks were based on ideas of supervised learning \cite{pomerleau1991efficient}. Not long after, problems such as data scarcity \cite{natarajan2013accelerating} and covariate shifts \cite{sugiyama2012machine} forced the adoption of fresh ideas from the field of Reinforcement Learning (RL).

Generally speaking, RL based methods held the promise of addressing the fundamental limitation of supervised approaches. That is, accounting for potentially disastrous approximation errors: $||\hat{a}-a^*||$, by observing their effect throughout complete trajectories.\\
%
%
For example, work such as Forward Training \cite{ross2010efficient}, and SMILe \cite{ross2010efficient} offered gradual schemes that learn different policies for each time-step. At time $t$, Forward Training will try to compensate for drifting errors by training a policy $\pi_t$ on the {\em actual} state-distribution induced by policies $\pi_0$ to $\pi_{t-1}$. SMILe, on the other hand, will repeatedly query the expert on the actual trajectories visited by the agent until convergence. However, both approaches, and counterparts of their kind were not suitable for challenging imitation setups. Tasks with a long time horizon were ruled out because of the requirement to train a policy for each time-step. Real world problems were excluded because they couldn't provide an expert-on-demand as the methods require.

\subsection{Imitation Learning and No-Regret Algorithms}
Soon after, \citet{ross2011reduction} introduced the DAgger algorithm. While similar in spirit to SMILe, DAgger operates in a slightly different manner. It proceeds by gathering experience using an agent-expert mixed behavioral policy: $$\pi_t=\beta(t)\pi^* + \big(1- \beta(t)\big)\hat{\pi_t},$$ where $\pi^*, \pi$ and $\hat{\pi}$ are the expert, behavior and agent policies respectively. The data $\mathcal{D}_{t}$ that was gathered using $\pi_t$ is aggregated together with all datasets collected up to time $t$: $\mathcal{D}_{t} = \bigcup_{\tau=0}^t \mathcal{D}_\tau$. Eventually, a policy is trained on the cumulative set $\mathcal{D}_{t}$ that is labeled with the help of the expert: $$\hat{\pi}_{t+1} = \argmin_{\pi} \sum_{s_i \in \mathcal{D}_t} \ell(\pi(s_i), \pi^*(s_i))$$
Explained differently, at each iteration, DAgger is training a policy to succeed on all trajectories seen so far. This is in contrary to most RL approaches that wish to fit fresh online data only. With this trait in mind, DAgger can be thought of as an online no-regret imitation algorithm. Moreover, it is shown by the authors that no other imitation algorithm can achieve better regret.

\subsection{Adversarial Networks and Imitation Learning}
A significant breakthrough in the field occurred at 2016 with the introduction of Generative Adversarial Imitation Learning (GAIL) \cite{ho2016generative}, an imitation algorithm closely related to the celebrated GAN architecture. GAN was originally presented as a method to learn generative models by defining a two-player zero-sum game:
\begin{equation}
\begin{split}
\label{eq:GAN}
\arg&\min_{G} \, \argmax_{D \in (0,1)} \quad  \\ &\mathbb{E}_{x \backsim p_E} [\log D(x)] +
\mathbb{E}_{z \backsim p_z} \big[ \log \big( 1-D(G(z)) \big) \big],
\end{split}
\end{equation}

where $p_z$ is some noise distribution, player $G$ represents the generative model and $D$ is the judge. GAIL showed that GAN fits imitation problems like a glove. By modifying $G: \eta \rightarrow x$ to represent a policy $\pi: s \rightarrow a$, GAIL showed how to harness GAN for imitation purposes:
\begin{equation}
\begin{split}
\label{eq:GAIL}
\arg&\min_{\pi} \, \argmax_ {D \in (0,1)} \, \\ & \mathbb{E}_{\pi}[\log D(s,a)] + \mathbb{E}_{\pi_E} [\log(1-D(s,a))] -\lambda H(\pi),
\end{split}
\end{equation}

The motivation behind GAIL, i.e., to use GAN for imitation, is to rely on a neural network to build a dynamic decision rule to classify between the expert's and the agent's state-action pairs. GAIL uses the continuous classification score as a proxy reward signal to train the player. Other than that, GAIL also proved to be efficient with respect to the number of expert examples it requires. This is partially explained by the fact that even though expert examples are scarce, the algorithm enjoys an unlimited access to agent examples through simulation. Loosely speaking, having infinitely many agent examples allow the discriminative net to gain a precise understanding of its behavior, and as a result to also understand it’s differences from the expert.\\
However, while GAIL is efficient in terms of expert examples, this is clearly not the case regarding the required number of environment interactions. The high sample complexity is explained by the fact that GAIL's update rule is based on the famous REINFORCE algorithm \cite{williams1992simple}. REINFORCE offers an approximation to the true gradient, and is primarily used in situations where it is hard, or even impossible to calculate the original gradient. However, REINFORCE suffers from a high variance and is known to require numerous iterations before converging.
\subsection{Model-Based Adversarial Imitation Learning}
To compensate for this inefficiency, model-based Imitation Learning (mGAIL) \cite{baram2016model}  offered to take an opposite approach. While GAIL applies {\em gradient approximations} on samples obtained from the {\em original environment}, mGAIL applies the {\em original gradient} on samples obtained from an {\em approximated environment}.\\
Put differently, mGAIL offers a model-based alternative that attempts to learn a differentiable parametrization of the forward model (transition function). Using this approach, a multi-step interaction with the environment creates an end-to-end differentiable graph that allows to backpropagate gradients through time, thus, enabling to calculate the original gradient of the objective function. mGAIL's advantage of using the original gradient comes at the cost of learning an accurate forward model, a task that often proves to be extremely challenging. Errors in the forward model can bias the gradient up to a level where convergence is again at risk.
\subsection{The limitation of Current Approaches}
While GAIL and mGAIL complement each other, both methods amount to a standard imitation setup that requires a shared action space between the expert and the agent.
To understand why, it is enough to revisit GAIL's decision rule (Eq~\ref{eq:GAIL}).
In both methods, the rule is based on the joint distribution of states and actions.
Not only that such a decision rule is limiting the agent to operate in the expert’s action domain, but it adds further optimization complications. State and actions are completely different quantities and embedding them in a joint space is not trivial.\\
In this work, we argue that the right decision rule should not be based on the joint state-action distribution $p(s,a)$, but rather on the state-transition distribution $p(s_{t-1} \rightarrow s_t)$. Using the state-transition distribution we can neutralize the presence of the agent performing the task and instead focus on imitating the effects it induces on the environment.\\
\subsection{PbRL and Adversarial Imitation Learning}
Even though adversarial-based methods have proved successful for imitation, algorithms of this type suffer from an acute problem: they induce a non-stationary MDP. The reward, which is derived from a continually-adapting classification rule, is constantly changing. As a result, estimation of long-term returns, an underlying ingredient in most RL algorithms, becomes almost infeasible \cite{nareyek2003choosing}, \cite{koulouriotis2008reinforcement}. We believe that alleviating this problem is possible using the concept of PbRL.\\
The motivation behind PbRL is to alleviate the difficulty of designing reward functions. Originally, PbRL was designed as a paradigm for learning from non-numerical feedback \cite{furnkranz2011preference}. Instead, PbRL tries to train agents using preferences between states, actions or trajectories. The goal of the agent in PbRL is to find a policy $\pi^*$ that maximally complies with a set of preferences $\zeta$. Assume two trajectories $\tau_1, \tau_2$. A preference $\tau_1 \succ \tau_2 \in \zeta$ is satisfied if: $$\tau_1 \succ \tau_2 \iff Pr_\pi (\tau_1) > Pr_\pi (\tau_2) .$$
Generally speaking, PbRL methods are divided into three categories. The first includes algorithms that directly search for policies that comply with expert preferences \cite{wilson2012bayesian}. The second consists of "model-based" approaches that rely on a preference model \cite{furnkranz2012preference}. And the third encompasses methods that try to estimate a surrogate utility $U(\tau)$ for a given trajectory \cite{akrour2011preference}. We refer the reader to \citet{wirth2017survey} for a more comprehensive survey on PbRL.


\section{Algorithm}
As of today, the prevalent approach to imitation is to take a GAN-like approach. To understand this connection better, we recall that GANs in the context of imitation, or RL in general, can be best understood as a form of an actor-critic architecture \cite{pfau2016connecting}. Therefore, in the following section, we present a family of advantage actor-critic algorithms for Inspiration learning tasks.\\
\subsection{Actor Critic Methods}
In an actor-critic algorithm, one of the prevailing approaches in reinforcement learning \cite{konda2000actor}, one maintains a separate parameterization for the policy (actor) and the state-value function (critic).
The role of the actor is straightforward- to represent the policy. The role of the critic, on the other hand, is to assess the expected performance of the policy based on experience. The critic’s estimation is used as a baseline to determine whether the current behavior should be strengthened (if better than the baseline), or weakened (if worse). Numerous actor-critic variations exist \cite{vamvoudakis2010online}, \cite{bhasin2013novel}. Among the common ones is the advantage actor-critic Architecture \cite{peters2008natural}.

\subsection{Advantage Actor-Critic for Inspiration Learning}
The advantage function of a state-action pair $(s,a)$ is defined as $A(s,a) = Q(s,a)-V(s)$.
The function comprises of two parts: an action-dependent term $Q(s,a)$, and an action-independent term $V(s)$. Because of their structure, advantage functions are commonly used to score gradients in policy gradient algorithms. $Q(s,a)$ is often approximated by the return from online rollouts: $R^\pi(s,a)=\sum_{t=0}^T{\big[\gamma^t \cdot r_t | s_0=s,a_0=a, a_t \sim \pi\big]}$, while $v(s)$ is trained to predict the expected discounted return of states through regression. The approximated advantage function is given by:
\begin{equation}
\label{advantage}
A(s,a) \approx R(s,a)-V(s)
\end{equation}
However, as in any imitation challenge, the reward signal is absent. In the following, we describe how using PbRL we are able to synthesis robust classification-based rewards that can be integrated into any advantage actor-critic algorithm. We start by describing the proposed rewards.\\

\begin{itemize}
	\item \textbf{Basic scoring:} The first variant we consider score actions by considering their raw classification score. It is simply given by:
	\begin{equation}
	\label{basic_reward}
r^{BASIC}(a_t,s_t)=c_a(s_t)
\end{equation}
We note that $c_a(s_t)$ is the classification score for the $s_t\rightarrow s^a_{t+1}$ transition, where $s^a_{t+1}$ is the state following $s_t$ when choosing action $a$.
	
	\begin{figure*}[ht]
		\begin{subfigure}{.295\textwidth}
			\centering
			Breakout
			\includegraphics[ width=\linewidth]{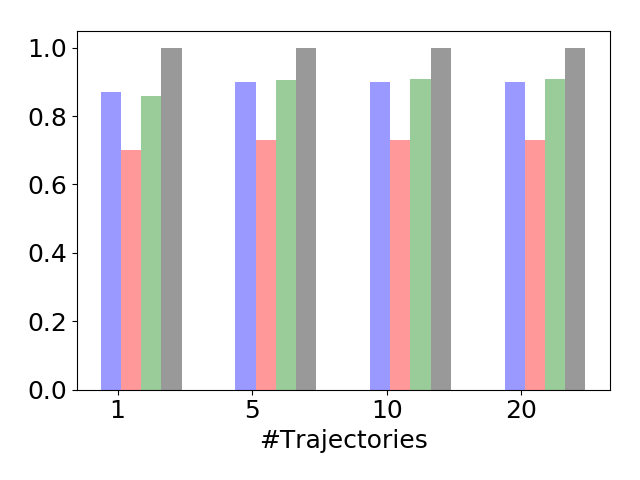}
			\caption{}
		\end{subfigure}\hfill
		\begin{subfigure}{.295\textwidth}
			\centering
			Enduro
			\includegraphics[ width=\linewidth]{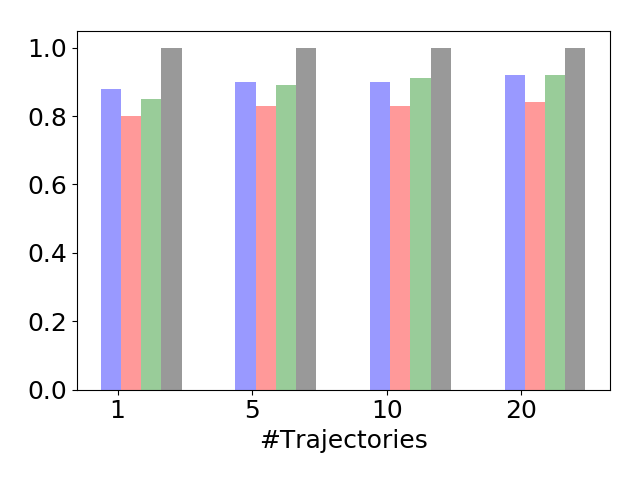}
			\caption{}
		\end{subfigure}
		\begin{subfigure}{.295\textwidth}
			\centering
			Seaquest
			\includegraphics[ width=\linewidth]{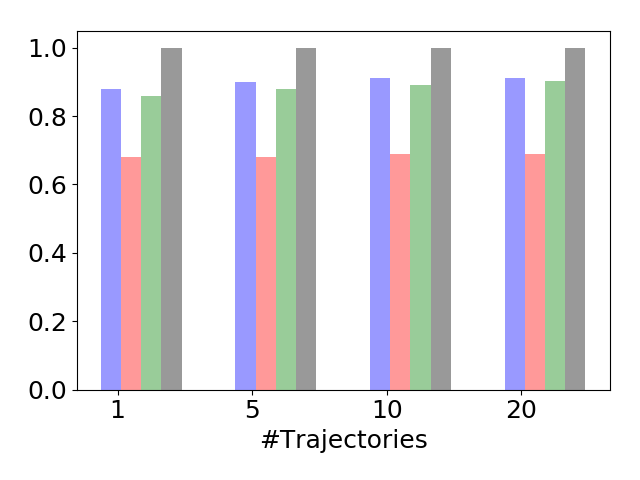}
			\caption{}
		\end{subfigure}\hfill
			\begin{subfigure}{.1\textwidth}
		\centering
		\includegraphics[ width=\linewidth]{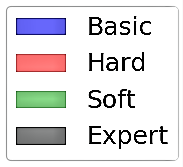}
	\end{subfigure}\hfill
		\caption{\textbf{Shared Action Imitation:} results for using our proposed method in the standard imitation framework where the expert and the agent share the same action space. We run experiments on three Atari2600 games: Breakout (a), Enduro (b) and Seaquest (c)}
		\label{fig:shared_action}
	\end{figure*}
		
\item \textbf{Preferential scoring:} Working with a discrimination based reward pose challenges that arise from its non-stationary nature. To facilitate this problem, we suggest applying a ranking transformation $\mathcal{T}$ on the raw classification scores. $$\mathcal{T} \circ c_a(s_t)=\frac{\big|\big\{\tilde{a} \in \mathcal{A} \,| \, c_{\tilde{a}}(s_t)\le c_a(s_t)\big\}\big|}{|\mathcal{A}|}$$
Put in words, $\mathcal{T}$ calculates action-based discrimination preferences. Doing so, we are able to discard nuance in temporary classification scores, and encourage stationarity through the course of training. The preferential reward is given by:
\begin{equation}
	\label{preferential_reward}
	r^{PRE}(a_t,s_t)=\mathcal{T} \circ c_a(s_t)
\end{equation}
\item \textbf{soft-Preferential scoring:} In cases where classification scores are approximately the same: $c_{a_i(s_t)} \approx c_{a_j(s_t)}$, a hard ranking transformation is likely to deform the  decision rule to a high degree. Relaxing the deformation can be carried out in several ways. In this paper, we choose to apply a \textit{softmax} transformation to the raw classification scores:
$$\mathcal{T}_s \circ c_a(s_t)=\frac{e^{c_a(s_t)}}{\sum_{\tilde{a}}{e^{c_{\tilde{a}}(s_t)}}} \, .$$
The soft preferential reward is therefore given by:
\begin{equation}
\label{smooth_preferential_reward}
r^{sPRE}(a_t,s_t)= \mathcal{T}_{s}\circ c_a(s_t)
\end{equation}
\end{itemize}

The algorithm we propose requires a set of expert trajectories that include states only $\tau_E=\{s_0^*,s_1^*,s_2^*,...,s_T^*\}$. At each iteration, the agent gathers experience according to its current policy. At the same time, classification scores $c_{a_t}(s_t)$ and state-value estimations $V(s_t)$ are recorded at each time step $t$. We then derive a reward function as explained above. For efficiency, the policy improvement step proceeds from end to start: an advantage function $A(a_t,s_t)$ is approximated using the returns $R(a_t,s_t)$ and the value estimations $V(s_t)$. Finally, a policy gradient step can take place: $$\theta_{\pi} \leftarrow \theta_\pi + \eta \nabla_{\theta_{\pi}} \log
\pi(a_t|s_t:s_{t-k})A(a_t,s_t)$$
The full algorithm is presented in Algorithm~\ref{algorithm}.

\begin{algorithm}[]
	\caption{\textbf{Advantage Actor-Critic for Inspiration Learning - pseudo-code for a basic reward function}}
	\begin{algorithmic}[1]
		\label{algorithm}
		\STATE \textbf{Input:} \\
		\begin{itemize}
			\item Expert trajectories $\tau_E=\{s_0^*,s_1^*,s_2^*,...,s_T^*\}$
			\item Actor parameterization $\theta_\pi$
			\item value function parameterization $\theta_v$
			\item expert-agent classifier parameterization $\theta_c$
			\item Mini-trajectory length $L$
			\item State stack parameter $k$
			\item Learning rate $\eta$
			\item Max number of steps $T_{max}$
		\end{itemize}
		\STATE \textbf{repeat}
		\STATE \-\hspace{0.5cm} Reset gradients: $d\theta_\pi \leftarrow 0, d\theta_v \leftarrow 0, d\theta_c \leftarrow 0$
		\STATE \-\hspace{0.5cm} Reset episode counter $t \leftarrow 0$
		\STATE \-\hspace{0.5cm} \textbf{repeat}
		\STATE \-\hspace{1.0cm} Sample action $a_t \sim \pi(a|s_t)$
		\STATE \-\hspace{1.0cm} Expert/agent classification $c_t = c_{a_t}(s_t)$
		\STATE \-\hspace{1.0cm} Value estimation $v_t=v(s_t)$
		\STATE \-\hspace{1.0cm} Interact $s_{t+1} = $ Env$(a_t,s_t)$ 
		\STATE \-\hspace{1.0cm} Update episode counter $t \leftarrow t + 1$
		\STATE \-\hspace{1.0cm} Update total step counter $T \leftarrow T + 1$
		\STATE \-\hspace{0.5cm} \textbf{while} $t<L$ or $done$
		\STATE \-\hspace{0.5cm} $
				\quad A=\begin{cases}
			             0, \quad $if reached a terminal state$\\
         			     c_{a_t}(s_t), \quad $else$\\
		\end{cases}$
		\STATE \-\hspace{0.5cm} \textbf{for} $i=t$ \textbf{to} $0$ \textbf{do}
		\STATE \-\hspace{1.0cm} $A \leftarrow c_{a_t}(s_t) + \gamma A$
		\STATE \-\hspace{1.0cm} $\theta_\pi: d\theta_\pi \leftarrow  d\theta_{\pi} + \eta  \frac{\partial}{\partial \theta_{\pi}} \log \pi(a_i|s_t) \, A$
		\STATE \-\hspace{1.0cm} $\theta_v: d\theta_v \leftarrow d\theta_v - \eta\frac{\partial}{\partial \theta_v} \big[A-v(s_t;\theta_v)\big]^2$
		\STATE \-\hspace{1.0cm} $\theta_c: d\theta_c \leftarrow d\theta_c - \eta\frac{\partial}{\partial c} \log \big[ c_{a_t}(s_t;\theta_c) \big]$
		\STATE \-\hspace{1.0cm} $\theta_c: d\theta_c \leftarrow d\theta_c - \eta\frac{\partial}{\partial c} \log \big[1-c_{a^*}(s_t^*;\theta_c) \big]$
		\STATE \textbf{while} $T<T_{max}$
	\end{algorithmic}
\end{algorithm}


\section{Empirical Evaluation}
In this section, we assess the performance of our actor-critic algorithms in action. We used a standard shared parameterization for the three components of our algorithm: the actor, the state-value function and the expert-agent classifier. Our implementation uses a convolutional neural network with two layers of 64 convolution channels each (with rectified linear units in between), followed by a 512-wide fully connected layer (with rectified linear unit activation), and three output layers for $\pi$, $v$ and $c$. Unless otherwise mentioned, states include a stack of last $k=4$ frames, and experts were trained using vanilla advantage-actor-critic algorithm \cite{1606.01540}. We use $N=\{1,5,10,20\}$ expert trajectories across all experiments\footnote{All experiments were conducted on a single GPU machine with a single actor.}.
\subsection{Shared Actions Imitation: }
The purpose of the first set of experiments is to test our algorithm in the familiar imitation learning setup: an expert and an agent that share the same action space. We recall that only expert states are recorded, and the agent has no access to any ground truth actions whatsoever. We tested three Atari2600 games: Breakout, Enduro and Seaquest. Results for this section are shown in Figure~\ref{fig:shared_action}.

\subsection{Continuous to Discrete Imitation: }
In the second set of experiments, we wish to test our method in a setup where the expert is using a continuous action space and the agent uses a discrete one. To test this setup we use the following two environments:
\subsubsection{Roundabout Merging Problem:}
A car is merging into a roundabout that contains two types of drivers: aggressive ones that do not give right of way when a new car wishes to enter the roundabout, and courteous drivers that will let the new car merge in. The car is positively rewarded after a successful merge and negatively rewarded otherwise (we note that success is defined over a set of considerations including efficiency, safety, comfort etc.). We use a rule-based logic for the expert in this case. The expert is responsible to output two continuous signals: 1) a one-dimensional steering command $a_{steer} \in \mathbb{R}$ and 2) a one-dimensional acceleration command $a_{accel} \in \mathbb{R}$. The agent, on the other hand, is allowed to choose from a discrete set of six commands indicating longitudinal $dx$ and transverse $dz$ shifts from its current position $x_t$. Results are shown in Figure~\ref{fig:cont2disc}.
\subsubsection{Hopper-v2: }
\begin{figure*}
	\begin{subfigure}{.48\textwidth}
		\centering
		Roundabout
		\includegraphics[ width=\linewidth]{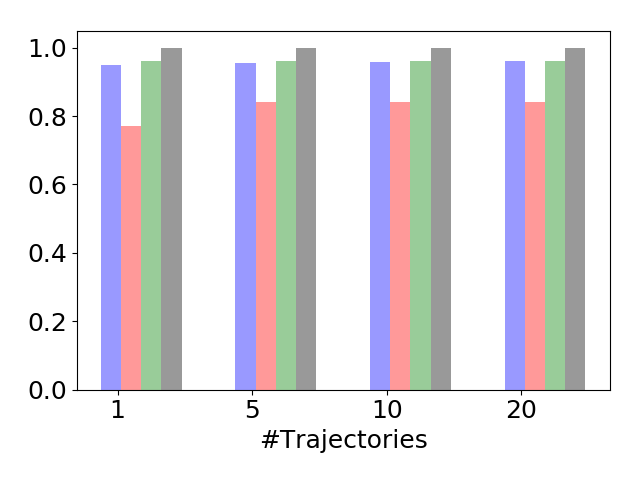}
		\caption{}
	\end{subfigure}\hfill
	\begin{subfigure}{.48\textwidth}
		\centering
		Hopper-v2
		\includegraphics[ width=\linewidth]{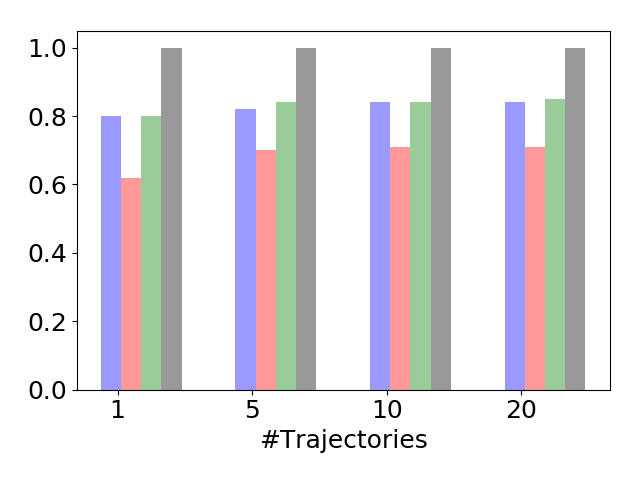}
		\caption{}		
	\end{subfigure}
	\caption{\textbf{Continuous to Discrete Imitation:} results for using our method to imitate a continuous teacher using a discrete agent. For both tasks, we extracted a set of 7 actions by clustering the set of expert actions. On the left, we can see how such a discrete imitator can learn competent policy for the roundabout merging task. On the right, we see results on the MuJoCo benchmark.}
	\label{fig:cont2disc}
\end{figure*}

A continuous control task modeled by the MuJoCo physics simulator \cite{todorov2012mujoco}. The expert uses a continuous action space to balance a multiple degrees of freedom monoped to keep it from falling. We use the Trust Region Policy Optimization \cite{schulman2015trust} algorithm to train the expert policy. We present the agent with the expert demonstrations and restrict him to use a set of 7 actions only (obtained by applying K-means clustering \cite{Agarwal:2004:KMP:1055558.1055581} on the expert actions). Results for this experiment are presented in Figure~\ref{fig:cont2disc}.

\subsection{Skills to Primitives Imitation: }
\begin{figure*}
	\begin{subfigure}{.33\textwidth}
		\centering
		Breakout
		\includegraphics[ width=\linewidth]{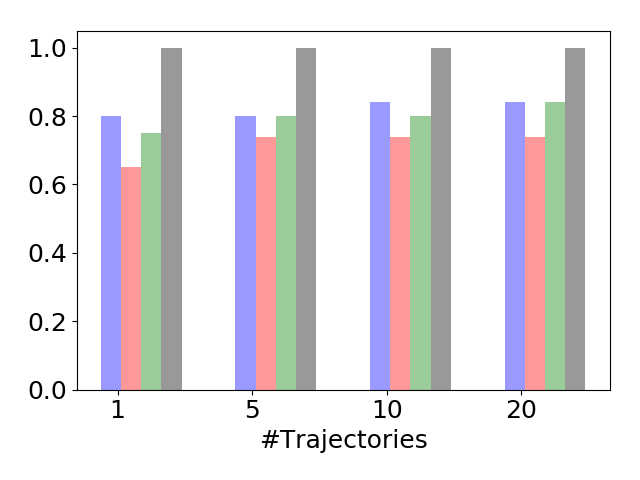}
		\caption{}
	\end{subfigure}\hfill
	\begin{subfigure}{.33\textwidth}
		\centering
		Enduro
		\includegraphics[ width=\linewidth]{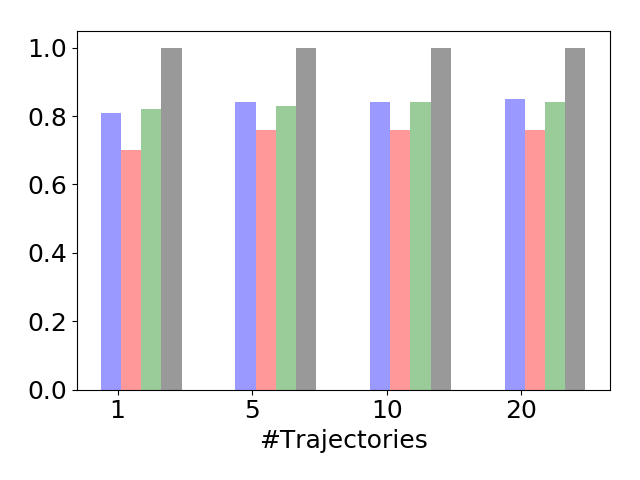}
		\caption{}
	\end{subfigure}
	\begin{subfigure}{.33\textwidth}
		\centering
		Seaquest
		\includegraphics[ width=\linewidth]{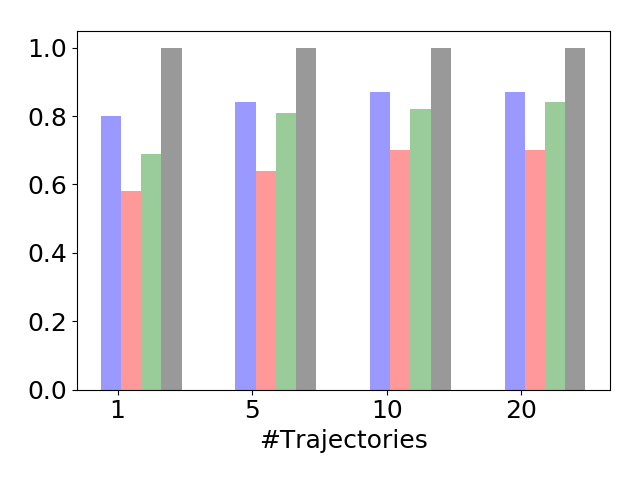}
		\caption{}
	\end{subfigure}\hfill
	\caption{\textbf{Skills to Primitives imitation:} results for using our method to train a low-level agent (that uses primitive actions) to imitate a high-level teacher (that plans using skills). In this experiment, we trained the expert using a set of handcrafted skills. Afterward, we used our algorithm to train an agent that is allowed to use primitive actions only. Not surprisingly, we obtain similar performance as in the shared-action space setup, since the agent is agnostic to the structure of the expert's policy and is only exposed to state-transitions.}
	\label{fig:macro2prim}
\end{figure*}

In the third set of experiments, an expert is first trained to solve a task using a set of predefined skills \cite{sutton1999between}. We then trained the agent, equipped with primitive actions only, to imitate the expert. We note that in this experiment, expert skills did not include primitive actions that are not available to the agent. We tested the same Atari games (Breakout, Enduro and Seaquest). The agent was able to achieve expert performance in all games (see Figure~\ref{fig:macro2prim}).

\subsection{Primitives to Skills Imitation:}
\begin{figure*}
	\begin{subfigure}{.48\textwidth}
		\centering
		Breakout
		\includegraphics[ width=\linewidth]{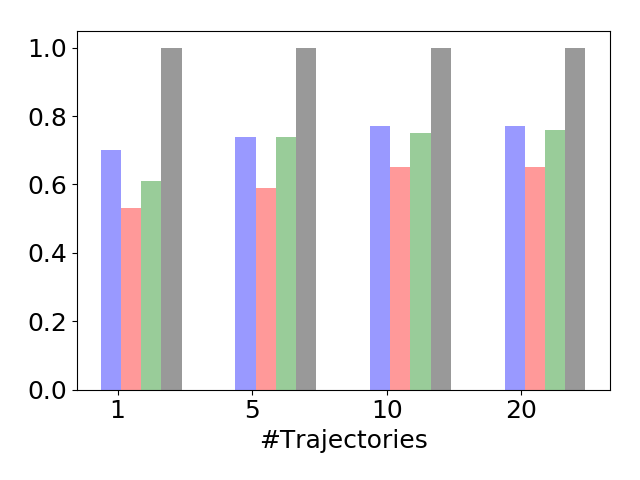}
		\caption{}
	\end{subfigure}\hfill
	\begin{subfigure}{.48\textwidth}
		\centering
		Enduro
		\includegraphics[ width=\linewidth]{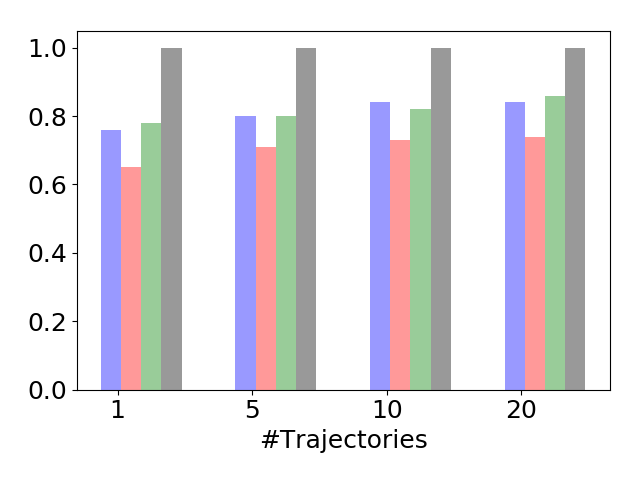}
		\caption{}
	\end{subfigure}
	\caption{\textbf{Primitives to Skills imitation:} results for using our method to train a hierarchical agent using demonstrations from a flat expert. On the contrary to the reversed setup (skills to primitives), we observe a slightly worse performance with respect to the shared-action space setup. We attribute this degradation to a model-mismatching between the handcrafted skills and the ground truth actions.}
	\label{fig:prim2macro}
\end{figure*}

In the last set of experiments, an expert is trained using primitive actions only, while the agent is equipped with skills. As before, the skills did not include primitive actions not available to the expert. As before, we used predefined skills. We note that using our algorithm to learn the options themselves can prove to be a useful method to extract expert skills from demonstrations. However, this is out of the scope of this work and is a subject of further research. Results for this section are provided in Figure~\ref{fig:prim2macro}).

\section{Conclusion}
In this work, we show how imitation between differently acting agents is possible.
Our novelty lies in the observation that imitation is attained when two agents induce the same effect on the environment and not necessarily when they share the same policy.
We accompany our observation with a family of actor-critic algorithms. An actor – to represent the training agent. A critic – to a) classify state transitions into two classes (agent/expert), from which a reward signal is derived to guide the actor, and b) learn the state-value function for bias reduction purposes. \\
We provide results for various types of imitation setups including shared action space imitation, continuous to discrete imitation and primitive to macro imitation. Some of the results are surprising. For example the ability to distill a continuous action space policy using discrete sets (we show examples where $|\mathcal{A}_{agent}|=7$). However, some of the results are less intriguing. For instance, the ability to decompose a macro level policy into a primitive-level one is almost trivial in our case.
The critic is oblivious to the performing agent and is only concerned with the induced effect on the environment. Thus, knowledge transfer between agents that operate in different action domains is possible.
\subsection{The Importance of a Shared Parametrization}
We have also experimented with separate parametrization architectures where the agent-expert classifier is modeled by a completely separate neural network (in oppose to a shared parametrization architecture, where a single neural network is used to output all three signals). We found that shared parametrization produces significantly better results. We hypothesize that a shared parametrization is important because the same features that are used to embed the expert (for the expert-agent classification task), are also used to sample actions from. 
\subsection{Where Do We Go From Here?}
Our method allows agents to find new strategies (besides the expert’s one) to solve a task. If this is the case, then it’s fair to consider agents (or strategies) that do not entirely cover the expert’s state distribution, but perhaps just a vital subset of states that are needed to reach the end goal. We believe that this is an interesting direction for further research.\\
The flexible nature of our framework that allows imitating an expert in various strategies, can be used to obtain super-expert performance. By super we mean, policies that are safer than the expert, more efficient, robust and so on. Our method can be integrated as a building block in an evolutionary algorithm that can help evolve robots that are optimal for specific imitation tasks.\\
A typical setup of our method requires two ingredients: a) expert examples (e.g. video recordings) and b) a simulator to train the agent. Although not tested, we require the expert states and the simulator states to be the same (i.e., to be generated by the same source). We speculate that this is crucial in order to prevent the critic from performing vain classification that is based on state appearance and not on the state dynamic, as we would like it to be. We hold that this limitation can be removed to allow imitation between different state distributions. Such an improvement can be carried out for example by redesigning the critic to solve two separate tasks: 1) appearance classification and 2) dynamic classification, and deriving the reward from the latter. This improvement is out of the scope of this paper and will be explored in a subsequent work. 

\bibliography{paper_bib}
\bibliographystyle{aaai}

\end{document}